\documentclass[graybox]{styles/svmult}

\usepackage{mathptmx}       
\usepackage{helvet}         
\usepackage{courier}        
\usepackage{type1cm}        
\usepackage{makeidx}         
\usepackage{graphicx}        
\usepackage{multicol}        
\usepackage[bottom]{footmisc}
\usepackage{caption}
\usepackage{subfig}
\usepackage{ circuitikz }
\usepackage{float}
\usepackage{mathtools}
\usepackage{siunitx}
\usepackage{arydshln}
\usepackage{amsfonts}

\makeindex             

\begin{document}

\title*{Machine Learning Optimized Approach for Parameter Selection in MESHFREE Simulations}
\titlerunning{MESHFREE Parameter Optimization}
\author{Banerjee, Paulami; Padmanabha, Mohan; Sanghavi, Chaitanya; Michel, Isabel and Gramsch, Simone}
\institute{Paulami Banerjee \at Fraunhofer ITWM, Kaiserslautern, \email{paulami.banerjee@itwm.fraunhofer.de}
\and Mohan Padmanabha \at Fraunhofer ITWM, Kaiserslautern, \email{mohan.padmanabha@itwm.fraunhofer.de}
\and Chaitanya Sanghavi \at Fraunhofer ITWM, Kaiserslautern, \email{chaitanya.sanghavi@itwm.fraunhofer.de}
\and Isabel Michel \at Fraunhofer ITWM, Kaiserslautern, \email{isabel.michel@itwm.fraunhofer.de}
\and Simone Gramsch \at Frankfurt University of Applied Sciences, Frankfurt am Main, \email{simone.gramsch@fb2.fra-uas.de}}

\maketitle

\abstract{Meshfree simulation methods are emerging as compelling alternatives to conventional mesh-based approaches, particularly in the fields of Computational Fluid Dynamics (CFD) and continuum mechanics. In this publication, we provide a comprehensive overview of our research combining Machine Learning (ML) and Fraunhofer's MESHFREE software (\url{www.meshfree.eu}), a powerful tool utilizing a numerical point cloud in a Generalized Finite Difference Method (GFDM). This tool enables the effective handling of complex flow domains, moving geometries, and free surfaces, while allowing users to finely tune local refinement and quality parameters for an optimal balance between computation time and results accuracy. However, manually determining the optimal parameter combination poses challenges, especially for less experienced users. We introduce a novel ML-optimized approach, using active learning, regression trees, and visualization on MESHFREE simulation data, demonstrating the impact of input combinations on results quality and computation time. This research contributes valuable insights into parameter optimization in meshfree simulations, enhancing accessibility and usability for a broader user base in scientific and engineering applications.}

\section{Introduction}
\label{sec:Intro}

Conventional numerical simulation tools use a mesh (grid) to solve partial differential equations (PDEs). These mesh-based methods need a mesh that fits the shape of the geometry of/around which the simulation is carried out. The time and effort for mesh generation increases with the complexity of the computational domain. Especially in fluid flow cases with a deformable domain or free surface flows, this process gets even more complicated as the domain changes with time and the mesh has to be adapted accordingly. Switching towards meshfree methods will help to reduce this tedious process of re-meshing.  

There are different meshfree methods that are not subject to this problem. In this chapter, we concentrate on a Generalized Finite Difference Method (GFDM) based tool developed by Fraunhofer ITWM and Fraunhofer SCAI called MESHFREE (beta version 11.3) \cite{2022MESHFREESoftware}. It employs a general continuum mechanics approach which allows to solve a wide range of applications. In the field of classical fluid dynamics, it has been used successfully for the water management of vehicles \cite{2021kuhnertCar}, the evaluation and design of Pelton turbines \cite{2019kuhnertFSI}, the prediction of solution mining processes \cite{2021michel}, or the transmission of aerosols \cite{2024LeithaeuserAerosols}. MESHFREE is also very well suited to reliably simulate multiphase flows such as below-boiling point evaporation \cite{2023LeeEvaporation} as well as above-boiling point vaporization \cite{2024suchdeLiquidVapor}. In addition, complex material models can be efficiently integrated, for example for injection molding \cite{2022veltmaatInjection}, metal cutting \cite{2021UhlmannWetMetalCutting}, or processes in soil mechanics \cite{2017MichelSoilMechanics}. In GFDM, the domain is discretized by a randomly distributed cloud of numerical points that are used to solve the PDEs. Typically, the point cloud moves with the flow according to a Lagrangian formulation. Additional key parameters beyond those typically encountered in conventional mesh-based solvers are necessary, specifically related to point cloud management. Familiarity with GFDM is essential to comprehend these key parameters. 

Recent research has showcased the potential of involving machine learning (ML) models in both mesh-based and meshfree simulations. In \cite{huang2021machine} and \cite{10.5555/3433701.3433712}, the authors explored the use of ML for optimal mesh generation and super-resolution of spatio-temporal solutions, respectively. A framework for learning mesh-based simulations using graph neural networks called MeshGraphNets was introduced in \cite{pfaff2020learning}. For meshfree methods, \cite{hamrani2023machine} and \cite{lefloch2021mesh} delved into using ML for surrogate modeling, with the former emphasizing the universal approximation property of supervised ML models and the latter introducing a transport-based meshfree approach. Building upon this foundation, \cite{zehnder2021ntopo} extended the application of ML models to employ implicit neural representations for meshfree topology optimization. Together, these studies emphasize the promising role of ML in improving both the efficiency as well as accuracy of both mesh-based and meshfree simulations.

The primary objective of this publication is to assist inexperienced users in configuring the necessary parameters for setting up and executing MESHFREE simulations effectively. In addition, using sound ML strategies, our trained model provides users with an estimated range in which they can expect the results -- including result quality and computation time. The parameter suggestions are tailored to achieve satisfactory results while optimizing computational resources, using active learning approaches. In order to investigate the influence of selected parameters on the simulation results, a simplified 3D case was used, which represents a laminar flow around a cylinder, as described in \cite{Schaefer1996}. The effect of the selected parameters on the computation time as well as the drag and lift coefficients were systematically recorded and employed as data for the use of informed ML\cite{von2021informed} strategies to effectively address both tasks. In the following sections, we will take a deeper look into the mathematics behind MESHFREE (Sections \ref{sec:Numerics} and \ref{sec:PhysicalModel}), the research question (Section \ref{sec:researchQuestion}), the considered use case (Section \ref{sec:UseCaseDescription}), the approach we took to solve it (Section \ref{sec:approach}), and the results we have obtained (Section \ref{sec:Results}). The concluding remarks are summarized in Section \ref{sec:Conclusion}.

\section{Numerics Based on GFDM} 
\label{sec:Numerics}
 
This section covers numerical point cloud management in MESHFREE, encompassing point distribution, neighborhood definition, and local refinement strategies.

\subsection{Point Cloud Management} 
\label{subsec:Numericalpoint}
The computational domain is discretized with numerical points (also known as point cloud) that are irregularly distributed in the interior and along the boundary. The generation and management of the points is comparatively easy and can be automated. An overview of different strategies can be found in \cite{2023suchdePointCloudGeneration}. A meshfree advancing front technique \cite{Lhner1998AnAF} is used for the point generation in MESHFREE.

The computational domain consists of $N = N(t)$ number of points with time $t$ and positions $\vec{x_i} \in \mathbb{R}^3$, $i = 1, ...,N$. The density/resolution of the point cloud is determined by the so-called interaction radius $h = h(\Vec{x},t) \in \mathbb{R}^+$, a sufficiently smooth function. Based on $h$, the local neighborhood $S_i$ of a point $\Vec{x_i}$ is defined as follows (see \cite{2021michel}, \cite{2019suchdefully}):
\begin{equation*} 
S_i = \{\vec{x}_j : ||\vec{x}_j -\vec{x}_i|| \leq h \}.
\end{equation*}
 
A sufficient quality of the point cloud has to be ensured locally throughout the simulation as the points move with the flow velocity according to the Lagrangian formulation. To avoid clustering, no neighboring points are allowed within a distance of $r_\mathrm{min} h$, where $r_\mathrm{min} \in (0,1)$. Additionally, to avoid large gaps there should be at least one neighboring point within a distance of $r_\mathrm{max} h$ with $r_\mathrm{max} \in (r_\mathrm{min},1)$. A typical choice is $r_\mathrm{min}=0.2$ and $r_\mathrm{max}=0.4$, resulting in 40 to 50 neighboring points in each interior neighborhood in 3D \cite{2017suchdeflux}. The number of neighboring points may be lower for points close to the boundaries of the domain \cite{2018suchdePointMovement}.

MESHFREE allows different strategies for local refinement with respect to the interaction radius $h$. The refinement can be geometry-dependent, where $h$ decreases with decreasing distance from the geometry (linear, spherical, or radial). Furthermore, there is also the possibility to define the interaction radius adaptively based on the simulation results such as velocity or pressure gradients. 

\subsection{Differential Operators}  
\label{subsec:DifferentialOperator}
MESHFREE employs a weighted least squares approximation approach to compute differential operators (see \cite{2021michel}, \cite{2018suchdeGFDMNS}). For example, the derivative approximation for a function $u$ at point $\Vec{x_i}$ is given by
\begin{equation*} 
\partial^* u (\vec{x}_i) \approx \widetilde{\partial}^*_i u = \sum_{j\in S_i} c^*_{ij} u_j,
\end{equation*}
where $*=x, y, z, \Delta,..$, $\partial^*$ is the continuous differential and $\widetilde{\partial}^*_i$ is the discrete differential at point $\vec{x}_i$. The stencil coefficients $c^*_{ij}$ are determined by a weighted least squares approximation that is independent of the function $u$. These coefficients are computed using a norm minimization process which ensures that monomials up to a specified order are differentiated exactly:
\begin{equation*}
    \min \sum_{j\in S_i} \left( \frac{c^*_{ij}}{W_{ij}} \right)^2,
\end{equation*}

where $W_{ij}$ is a weighting function with highest weight for the closest neighboring points. It is computed using the truncated Gaussian weighting function
\begin{equation*}
    W_{ij}= 
    \begin{dcases}
        \exp\left(-c_W \frac{||\vec{x}_j -\vec{x}_i||^2 }{h^2_i+h^2_j}\right),& \text{if } \vec{x}_j \in S_j\\
        0,  & \text{otherwise}
    \end{dcases}
\end{equation*}
for a constant $c_W \in [2,8]$. Smaller values of $c_W$ make the weighting function broader, allowing a bigger contribution of distant neighbors. Whereas, larger values lead to a stronger influence of the nearest neighbors.

\section{Physical Model}
\label{sec:PhysicalModel}

The physical model in MESHFREE is based on the conservation equations for mass, momentum, and energy in Lagrangian form. As temperature is irrelevant for the later use case in Section \ref{sec:UseCaseDescription}, we consider the following reduced set of equations:
\begin{align*}
    \frac{d \rho}{dt} &= -\rho \nabla^\mathrm{T}  \vec{u},  &
    \frac{d \vec{u}}{dt} &= - \frac{1}{\rho} \nabla p + \frac{1}{\rho}(\nabla^\mathrm{T} \mathbf{S})^\mathrm{T} + \vec{g}.
\end{align*}

Here, $\frac{d }{dt} = \frac{\partial }{\partial t} + \vec{u}^\mathrm{T} \nabla$ is the material derivative, $\nabla=\left(\frac{\partial}{\partial x}, \frac{\partial}{\partial y}, \frac{\partial}{\partial z}\right)^\mathrm{T}$ is the nabla operator, $\rho \in \mathbb{R}^+$ is the density, $\vec{u} \in \mathbb{R}^3$ is the velocity, $p \in \mathbb{R}$ is the pressure, $\vec{g} \in \mathbb{R}^3$ are the external forces, and $\mathbf{S} \in \mathbb{R}^3 \times \mathbb{R}^3$ is the stress tensor. For a detailed description of the time integration scheme used to solved these equations, see \cite{2018suchdeGFDMNS} and \cite{2018suchdePointMovement}.

With regards to the later use case of 3D flow around a cylinder, drag force $F_\mathrm{D}$ and lift force $F_\mathrm{L}$ on a surface $S$ are determined by
\begin{align*}
F_\mathrm{D} &= \int_{S} \left(\rho \nu \frac{\partial \vec{u}}{\partial \vec{n}} n_y - p n_x\right)dS, & F_\mathrm{L} &= -\int_{S} \left(\rho \nu \frac{\partial \vec{u}}{\partial \vec{n}} n_x + p n_y\right)dS, 
\end{align*}
where $\vec{n}=(n_x,n_y,n_z)^\mathrm{T}$ is the normal with respect to $S$ and $\nu \in \mathbb{R}^+$ is the kinematic viscosity. Drag coefficient $C_\mathrm{D}$ and lift coefficient $C_\mathrm{L}$ can be computed based on the mean velocity $\overline{u}$ as well as the diameter $D_\mathrm{cyl}$ and the length $H_\mathrm{cyl}$ of the cylinder:
\begin{align}
C_\mathrm{D} &= \frac{2\cdot F_\mathrm{D}}{\rho \bar{u}^2 D_\mathrm{cyl}H_\mathrm{cyl}}, & C_\mathrm{L} &= \frac{2 \cdot F_\mathrm{L}}{\rho \bar{u}^2 D_\mathrm{cyl}H_\mathrm{cyl}}. \label{eq:cDcL}
\end{align}

\section{Research Question}
\label{sec:researchQuestion}

Setting up and running a fluid flow simulation potentially involves tuning several parameters to obtain an accurate solution in optimized computation time, e.g. certain parameters might require domain knowledge. The more complex the application, the greater the number of parameters. Consequently, the balance between results accuracy and computation time is a major challenge, especially for inexperienced MESHFREE users. One option to assist inexperienced users is to pre-select those parameters that have a big influence and prescribe corresponding ranges with additional information about the trade-off between accuracy and efficiency.

\subsection{Parameter Selection}
\label{subsec:ParameterSelection}
Similar to other simulation tools, the characteristic length of the domain discretization is one of the most important parameters in MESHFREE, namely the interaction radius $h$. Here, we use a geometry-dependent local refinement with minimum interaction radius \emph{Hmin} close to the region of interest (cylinder in the later use case) and maximum interaction radius \emph{Hmax} outside of the region of interest with a linear increase from \emph{Hmin} to \emph{Hmax} in a transfer region. 

As described in Section 2, the GFDM approximation of differential operators at a given point is computed based on neighboring points within the local interaction radius $h$. The maximum number of neighboring points selected for the approximation can be controlled by the parameter \emph{max\_N\_stencil}, where the selection depends on the closest distance from the central point. 

Due to the Lagrangian formulation, point positions are updated in each time step. Consequently, the positions of the neighboring points will change as well, requiring an update of the local neighborhood for all the points. This is controlled by the parameter \emph{COMP\_DoOrganizeOnlyAfterHowManyCycles}.

The local solutions of pressure and velocity are computed with a fixed point iteration. Accordingly, convergence criteria have to be set by parameters \emph{eps\_p} and \emph{eps\_v}, respectively.

The approximation order of the local stencils $c^*_{ij}$ depends on the chosen order of monomials and can be controlled separately for the gradient operator (\emph{ord\_grad}), the Laplace operator (\emph{ord\_laplace}) and for the creation of new points (\emph{ord\_eval}).   

Furthermore, the constant $c_W$ in the weighting kernel $W_{ij}$  can be adapted for the Laplace operator (\emph{DIFFOP\_kernel\_Laplace}) as well as for the normal derivative (\emph{DIFFOP\_kernel\_Neumann}).  

Table \ref{tab:parameters} provides an overview of the selected parameters. More details regarding the respective parameters can be found in the MESHFREE documentation, see \cite{2024MESHFREEDocu}.

\begin{table}[H]
\caption{Parameters chosen for optimization and (range of) values}
\label{tab:parameters}
\begin{tabular}{p{3.4cm}p{6.2cm}p{1.8cm}}
\hline\noalign{\smallskip}
Parameter  & In MESHFREE & Values  \\ 
\noalign{\smallskip}\svhline\noalign{\smallskip}
Local interaction radius & \\ 
$\bullet$ minimum & \emph{Hmin} & 0.005 - 0.01 m \\ 
$\bullet$ maximum & \emph{Hmax} & 0.03 - 0.1 m\\
\noalign{\smallskip}\hline\noalign{\smallskip}
Neighbor points & & \\ 
$\bullet$ maximum number & \emph{max\_N\_stencil} & 30, 40\\
$\bullet$ update & \emph{COMP\_DoOrganizeOnlyAfterHowManyCycles} & 1 - 4\\
\noalign{\smallskip}\hline\noalign{\smallskip}
Convergence criteria & & \\ 
$\bullet$ velocity equation & \emph{eps\_v} & $1e^{-4}$,$1e^{-5}$,$1e^{-6}$ \\
$\bullet$ pressure equation & \emph{eps\_p} & $1e^{-4}$,$1e^{-5}$,$1e^{-6}$\\
\noalign{\smallskip}\hline\noalign{\smallskip}
Order of monomials & & \\ 
$\bullet$ gradient operator & \emph{ord\_grad} & 2, 3 \\
$\bullet$ Laplace operator & \emph{ord\_laplace} & 2, 2.9, 3\\
$\bullet$ creation of new points & \emph{ord\_eval} & 2, 3\\
\noalign{\smallskip}\hline\noalign{\smallskip}
Weighting kernel $c_W$ & & \\ 
$\bullet$ Laplace operator & \emph{DIFFOP\_kernel\_Laplace} & 2, 6 \\
$\bullet$ normal derivative & \emph{DIFFOP\_kernel\_Neumann} & 2, 5\\
\noalign{\smallskip}\hline\noalign{\smallskip}
\end{tabular}
\end{table}

\section{Use Case and Description of the Underlying Data} 
\label{sec:UseCaseDescription}

In this section, we outline the considered use case detailing the simulation setup -- including geometry, boundary conditions, and fluid properties. Additionally, we describe the data generation process featuring initial parameter ranges, sampling methodology, and an exploration of parameter correlations.

\subsection{Simulation Setup}
\label{subsec:GeometricSetup}
\begin{figure}[!ht]
\centering
\resizebox{0.6\textwidth}{!}{%
\begin{circuitikz}
\tikzstyle{every node}=[font=\Large]
\draw [, line width=0.8pt ] (8.5,16.75) rectangle (14.25,10.75);
\draw [, line width=0.8pt ] (23.5,19.5) rectangle (28,15);
\draw [line width=0.8pt, short] (14.25,16.75) -- (28,19.5);
\draw [line width=0.8pt, short] (8.5,16.75) -- (23.5,19.5);
\draw [line width=0.8pt, short] (14.25,10.75) -- (28,15);
\draw [dashed] (8.5,10.75) -- (23.5,15);
\draw  (16.5,14.5) ellipse (0.5cm and 0.75cm);
\draw [, dashed] (12.25,14.75) ellipse (0.5cm and 0.75cm);
\draw [short] (12.25,15.5) -- (16.5,15.25);
\draw [short] (12.25,14) -- (16.5,13.75);
\draw [<->, >=Stealth] (16.5,15.25) -- (16.5,17.25);
\draw [<->, >=Stealth] (14.25,10.5) -- (28,14.75);
\draw [<->, >=Stealth] (11.75,14.5) -- (8.5,13.75);
\draw [<->, >=Stealth] (12.25,14) -- (12.25,11.75);
\draw [<->, >=Stealth] (16,14.25) -- (17,14.75);
\draw [<->, >=Stealth] (8.25,10.5) -- (14.25,10.5);
\draw [<->, >=Stealth] (8.25,16.75) -- (8.25,10.75);
\node [font=\LARGE] at (12,11.25) {inflow};
\node [font=\Large] at (22.5,12.5) {2.5 m};
\node [font=\LARGE] at (25.75,16) {outflow};
\node [font=\large, color={rgb,255:red,150; green,150; blue,150}] at (21,16.5) {wall};
\node [font=\Large] at (7.25,13.75) {0.41 m};
\node [font=\Large] at (11.25,10) {0.41 m};
\node [font=\Large, color={rgb,255:red,120; green,120; blue,120}] at (14.5,14.5) {cylinder};
\node [font=\Large] at (17.75,14.25) {0.1 m};
\node [font=\Large] at (17.25,16) {0.16 m};
\node [font=\Large] at (10.25,15) {0.41 m};
\node [font=\Large] at (13.25,13) {0.15 m};
\node [font=\large, color={rgb,255:red,150; green,150; blue,150}] at (20,18.5) {wall};
\node [font=\large, color={rgb,255:red,150; green,150; blue,150}] at (21,13.5) {wall};
\node [font=\large, color={rgb,255:red,150; green,150; blue,150}] at (19.5,15.5) {wall};
\draw [->, >=Stealth] (26.5,12.25) -- (28.25,12.25);
\draw [->, >=Stealth] (26.5,12.25) -- (26.5,13.75);
\draw [->, >=Stealth] (26.5,12.25) -- (28.25,13);
\node [font=\Large] at (26.75,13.5) {Z};
\node [font=\Large] at (28,13.25) {X};
\node [font=\Large] at (28,11.75) {Y};
\end{circuitikz}
}
\caption{Geometrical setup of the 3D flow around a cylinder}
\label{fig:geometry}
\end{figure}
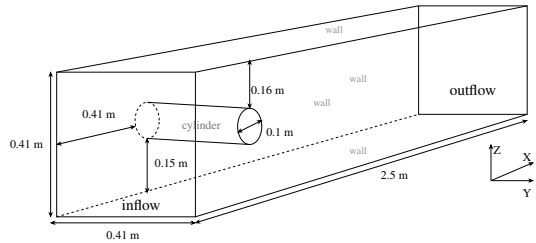

To study the influence of the selected parameters on accuracy and efficiency, a simple use case is employed. The 3D flow around a cylinder presented in \cite{Schaefer1996} was chosen due to its relative simplicity in setup and availability of validation data. Thereby, a 3D cylinder is placed in a hollow rectangular extruded domain perpendicular to the flow direction. The setup of the geometry is illustrated in Figure \ref{fig:geometry}. The length of the considered channel is $\SI{2.5}{\meter}$ along the $x$-axis. The height and width are $\SI{0.41}{\meter} \times \SI{0.41}{\meter}$ in $y$-$z$-plane. The cylinder extends along the entire width of the wall with a diameter of $D = \SI{0.1}{\meter}$. The cylinder is positioned $\SI{0.41}{\meter}$ downstream from the inflow and $\SI{0.15}{\meter}$ above the bottom wall. At the end of the channel, the flow exits through the outflow.

\begin{figure}[!ht]
    \centering
    \includegraphics[width=0.7\textwidth]{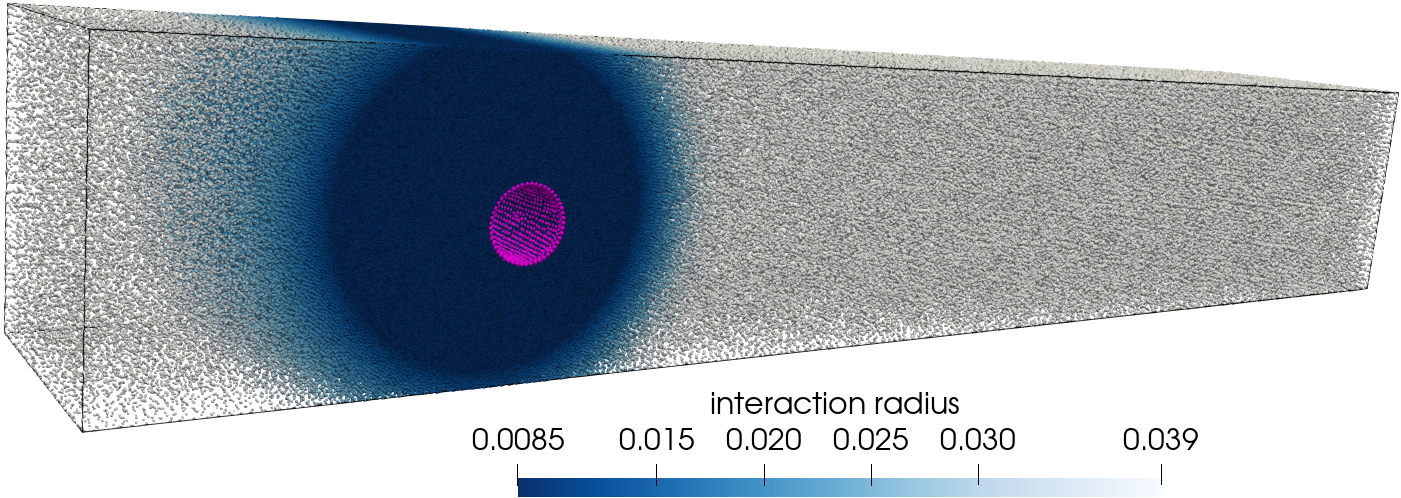}
    \caption{Point cloud and resulting interaction radius distribution for \emph{Hmin}$\ =\SI{8.5e-3}{\meter}$ and \emph{Hmax}$\ =\SI{3.9e-2}{\meter}$}
    \label{fig:hRefine}
\end{figure}

As described in Section \ref{subsec:ParameterSelection}, we employ a geometry-dependent definition of the interaction radius around the cylinder. To be more precise, this is a radial refinement in which the minimum interaction radius \emph{Hmin} is used up to a distance of $\SI{0.17}{\meter}$ from the cylinder. Outside of this, there is a linear increase towards the maximum interaction radius \emph{Hmax}. As an example, Figure \ref{fig:hRefine} shows the resulting point cloud for \emph{Hmin}$\ =\SI{8.5e-3}{\meter}$ and \emph{Hmax}$\ =\SI{3.9e-2}{\meter}$. The cylinder is colored magenta for better visibility.

For the simulations, we consider an incompressible Newtonian fluid with kinematic viscosity $\nu = \SI{1e-3}{\square\meter\per\second}$ and density $\rho = \SI{1.0}{\kilo\gram\per\cubic\meter}$. A homogeneous Dirichlet boundary condition is used at the channel walls and on the surface of the cylinder, i.e. $\vec{u} = \SI{0}{\meter\per\second}$. At the inflow, we prescribe a velocity profile in $x$-direction according to Equation \eqref{eq:VelocityProfile} and zero velocity in tangential directions, i.e. $u_y = u_z = \SI{0}{\meter\per\second}$:

\begin{equation}
    u_{x=0}(y,z,t) = \frac{16\cdot U_m y z (H_\mathrm{ch}-y)(H_\mathrm{ch}-z)}{H_\mathrm{ch}^4},
    \label{eq:VelocityProfile}
\end{equation}
where $U_m= \SI{2.25}{\meter\per\second}$ is the maximum inflow velocity, $H_\mathrm{ch}$ is the height of the channel, and $t = \SI{2}{\second}$ is the simulation time. Thus, the Reynolds number $Re$ is determined as
\begin{align*}
   Re &= \frac{\overline{u} D_\mathrm{cyl}}{\nu}, & \overline{u}&= \frac{4\cdot u_{x=0}\left(0.5\cdot H_\mathrm{ch}, 0.5\cdot H_\mathrm{ch}, t \right)}{9}.
\end{align*}

\begin{figure}[!ht]
    \centering
    \includegraphics[width=0.7\textwidth]{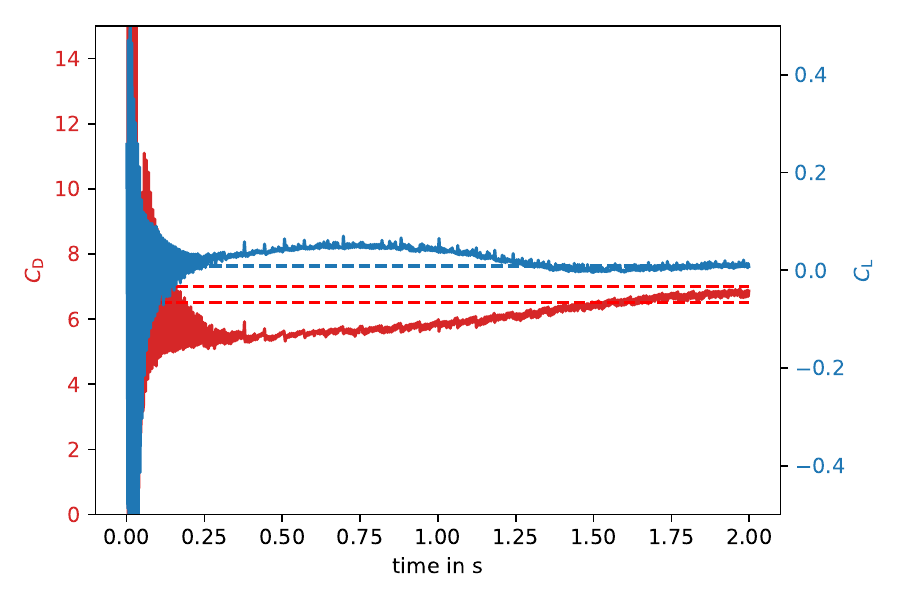}
    \caption{Comparison of the temporal evolution of drag (red line) and lift (blue line) coefficients for a selected data sample with the corresponding ranges for the benchmarks in \cite{Schaefer1996} (dashed lines)}
    \label{fig:DragLift}
\end{figure}

The drag and lift coefficients $C_\mathrm{D}$ and $C_\mathrm{L}$ are evaluated according to Equation \eqref{eq:cDcL} in each time step. To mitigate result fluctuations, the final drag and lift coefficients used in the informed ML strategies are averaged over the last 100 time steps. Figure \ref{fig:DragLift} shows an example of the evolution of $C_\mathrm{D}$ (red line) and $C_\mathrm{L}$ (blue line) over the simulation time for a selected data sample. The dashed lines represent the range of lift and drag coefficients for the benchmarks in \cite{Schaefer1996}. Good agreement with these values can be observed at the end of the MESHFREE simulation.


\subsection{Data Generation}
\label{DataGen}
The preliminary input parameter ranges were set as described in Table \ref{tab:parameters}. This served as the foundation for generating an initial sample space of 400 through Latin Hypercube Sampling (LHS) \cite{florian1992lhs}. LHS has been employed for its capacity to produce samples that closely mirror the true underlying distribution, resulting in smaller sample sizes compared to simple random sampling. This helped to enhance the efficiency in exploring the parameter space. Each data sample served as a specific input parameter combination for a MESHFREE simulation.

\subsection{Initial Data Exploration}
\label{DataExplore}
\begin{figure}[h!]
    \centering
    \includegraphics[width=\textwidth]{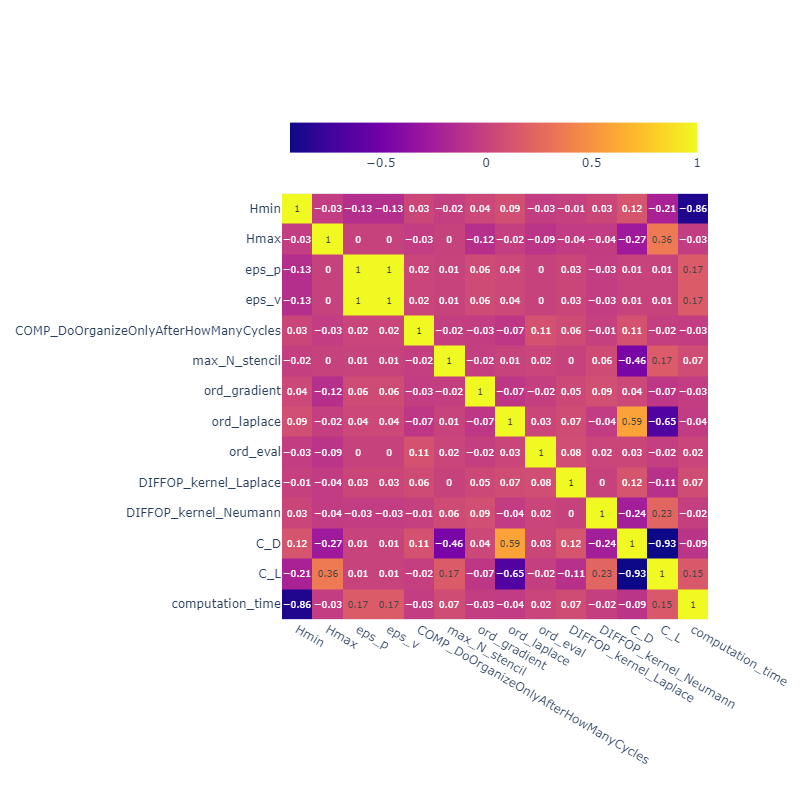}
    \caption{Correlation matrix for the selected parameters, drag and lift coefficients, as well as computation time}
    \label{fig:corr_heatmap}
\end{figure}

Figure \ref{fig:corr_heatmap} illustrates the correlation among the selected input parameters, simulation quality parameters like the drag and lift coefficients, as well as computation time using a heat map visualization, offering valuable insights into the relationships within the sampled space. As expected, the minimum local interaction radius \emph{Hmin} has a strong inverse effect on the computation time. Among the others, \emph{ord\_laplace} influences both the drag coefficient and the lift coefficient the most. The effects of individual input parameters on each of the output parameters have been illustrated in a dashboard\footnote{https://github.com/PaulamiBanerjee/ML4MESHFREE}. We summarize here a few of the takeaways derived from the plots available there.
\begin{itemize}
    \item The variance in both the drag coefficient and the lift coefficient increases with an increase in \emph{Hmax}.
    \item \emph{max\_N\_stencil} = \emph{40} produces better results for the drag coefficient.
    \item \emph{ord\_laplace} = \emph{2.9} produces better results for both the drag coefficient and the lift coefficient.
    \item Using \emph{DIFFOP\_kernel\_Neumann} = \emph{2} gives relatively better results for the lift coefficient.
\end{itemize}

Having gained these valuable insights from the initial data exploration, we now delve into the methods employed in our research to further investigate the influence of the selected parameters on accuracy and efficiency.

\section{Approach}
\label{sec:approach}

While much of the existing research involving the use of ML models in meshfree methods focuses on surrogate modeling -- aiming to comprehend the inherent workings of the numeric solver -- our objective is different. Rather than creating a surrogate model for the MESHFREE software, our aim is to enhance the user experience by assisting in selecting optimal parameter combinations to achieve a good balance between results accuracy and computation time, as already outlined in Section \ref{sec:researchQuestion}. Additionally, we have aspired to provide users with an estimated range instead of point estimates, within which the drag and lift coefficients of a particular simulation might reside, coupled with insights about the computation time. With these objectives in mind and acknowledging the limited data availability due to relying on simulations whose computation times range from hours to days (depending on the input parameter combination), we have adopted the following approach. Firstly, we implemented a parameter range selection strategy using Active Learning and secondly, we produced a conformal prediction of output parameter estimates using decision-tree-based ensemble regression models.

\subsection{Parameter Range Selection Using Active Learning}
\label{sec:AL}
\begin{figure}[h!]
    \centering
    \includegraphics[width=\textwidth]{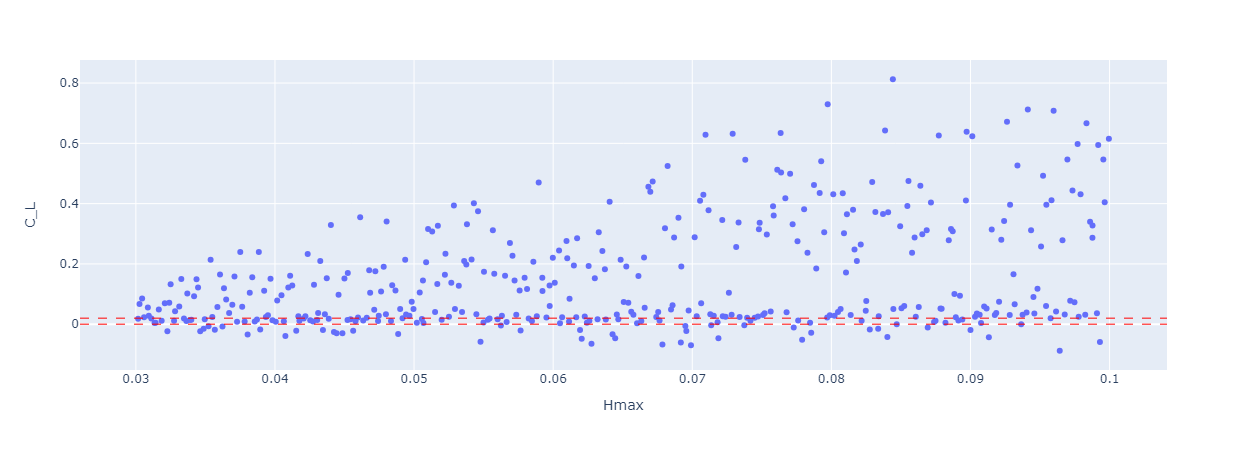}
    \caption{Cause-and-effect relationship between \emph{Hmax} and lift coefficient $C_\mathrm{L}$ (blue dots) and benchmark range in \cite{Schaefer1996} (red dashed line)}
    \label{fig:hmax-lift}
\end{figure}

While selecting values that yield favorable results is relatively straightforward in some instances (as highlighted in Section \ref{DataExplore}), it becomes more challenging in others. This complexity is exemplified in Figure \ref{fig:hmax-lift} illustrating the cause-and-effect relationship between \emph{Hmax} and the lift coefficient $C_\mathrm{L}$. As \emph{Hmax} decreases, the resulting lift coefficient are often within the red dashed lines, which indicate the benchmark range in \cite{Schaefer1996}. This corresponds to intuitive expectations. However, relying solely on visualization makes it difficult to pinpoint the precise optimal range for this specific parameter in combination with the other parameters.

In response to this challenge and recognizing the computational expense of running exhaustive simulations across all parameter combinations, we have turned to Active Learning strategies \cite{settles2009active}. Active Learning involves an adaptive learning algorithm that selectively identifies and labels the most informative samples, mitigating the demand for extensive labeled data. In our approach, the strategy entails training a Decision Tree classifier as the learning algorithm.

\begin{figure}[h!]
    \centering
    \includegraphics[width=0.9\textwidth]{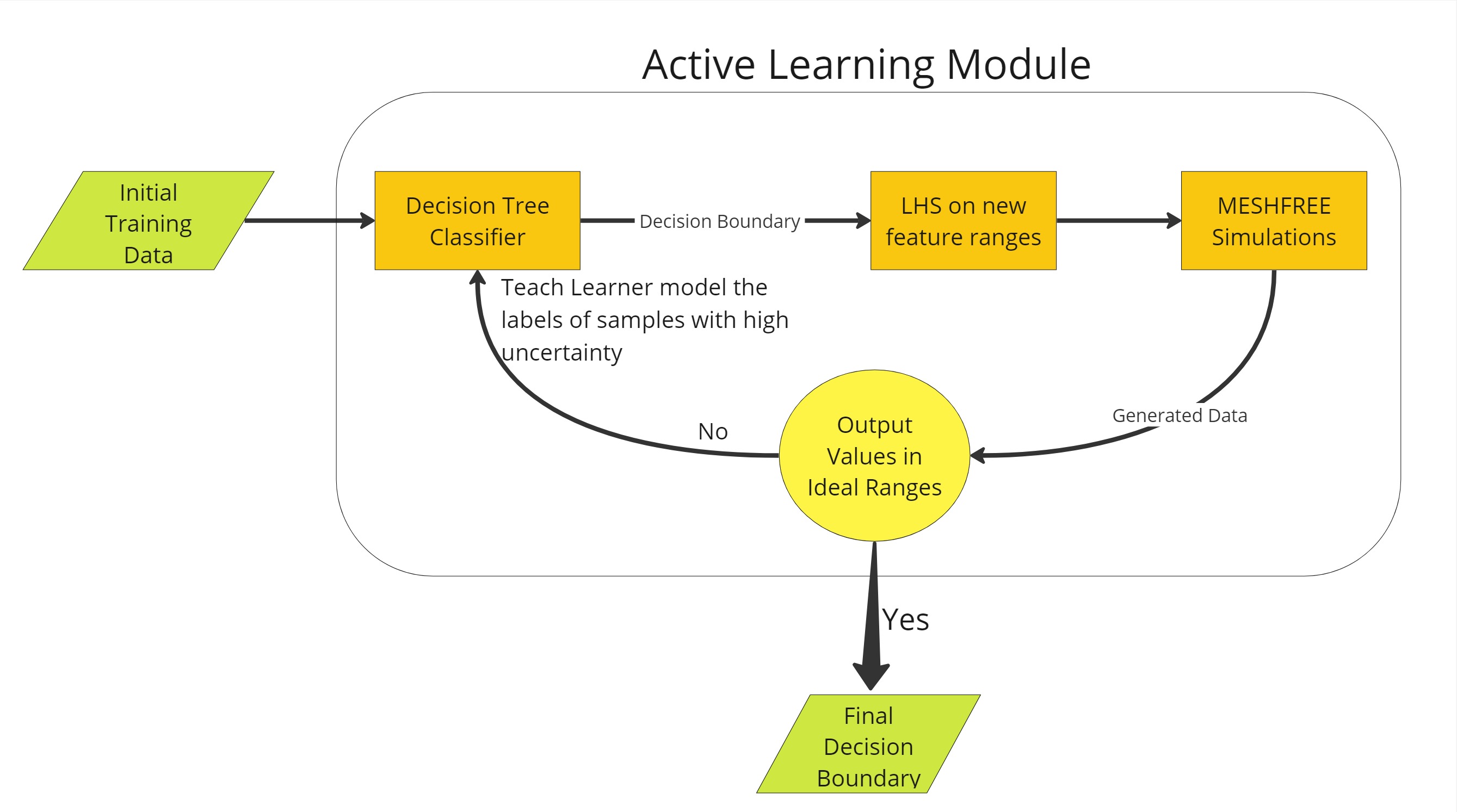}
    \caption{Flow of information in the Active Learning module}
    \label{fig:ALFlow}
\end{figure}

In each iteration of the learning process, we establish a decision boundary within the hypercube, where each dimension corresponds to a simulation parameter. Analogous to setting distinct zones within the hypercube, this decision boundary is particularly relevant in the context of our chosen regression model. It delineates regions in the decision trees where splits occur, guiding our focus during subsequent sampling. Feature ranges are then refined to include 10\% on either side of the decision boundary. This is done to accommodate the uncertainty of the learning process. LHS was performed within these refined feature ranges. MESHFREE simulations were conducted on the newly generated sample space, and the resulting outputs were labeled. If the resulting drag and lift coefficients do not lie within their respective optimum ranges, the learner tree is queried on this newly generated data about which cases it is more uncertain about, and the corresponding labels are taught to the learner. This iterative process was systematically repeated until we were able to successfully generate a whole set of data with all the output parameters in the optimum ranges. Figure \ref{fig:ALFlow} illustrates the flow of information in the described process.

\subsection{Conformal Prediction using Ensemble Regression Models}
Our objective is to forecast estimated intervals in which the three output parameters (drag and lift coefficient, computation time) are most likely to reside. To address this multi-output regression problem, we have implemented decision-tree-based ensemble regression models and the concept of Conformal Prediction \cite{taquet2022mapie} for predicting the estimated intervals. 

Tree-based regression models are well-suited for tabular data \cite{grinsztajn2022tree}. However, the ideal scenario for any machine learning task involves a substantial amount of data (ranging at least in tens of thousands). In our case, the reliance on computationally expensive simulations has resulted in a relatively limited dataset (ranging in hundreds). To address this challenge, boosting algorithms \cite{mayr2014evolution} were incorporated in the work as well. Boosting algorithms iteratively enhance the performance of a weak learner by giving more weight to instances with higher uncertainty, thereby creating a stronger ensemble model. An initial performance comparison of various regression models for each of the output parameters was conducted. The four best performing models in each case are presented in Table \ref{tab:InitPerfComp}. This comparison is based on the following error metrics: Mean Absolute Error (MAE), Mean Squared Error (MSE), Coefficient of determination (R2), and Mean Absolute Percentage Error (MAPE). Subsequently, three of the top performing models in each case -- Random Forest Regressor, Extra Trees Regressor, and Light Gradient Boosting (LGBM) Regressor -- were selected for further exploration.

\begin{table}[h!]
\centering
\caption{Initial performance comparison of regression models}
\label{tab:InitPerfComp}       
%
%
\begin{tabular}{p{0.9in}p{1.7in}p{0.4in}p{0.4in}p{0.4in}p{0.4in}}
\hline\noalign{\smallskip}
Output Parameter & Model & MAE & MSE & R2 & MAPE  \\
\noalign{\smallskip}\svhline\noalign{\smallskip}
Drag coefficient&Extra Trees Regressor & 0.7319  & 1.4253 & 0.9588 & 0.5416\\
 & Random Forest Regressor & 0.7344 & 1.4861 & 0.9541 & 0.4355\\
 & Light Gradient Boosting Machine & 1.0227  & 2.3034 & 0.9294 & 0.4572\\
 & Gradient Boosting Regressor & 1.0189  & 2.5714 & 0.9157 & 0.5386\\
\noalign{\smallskip}\hline\noalign{\smallskip}
Lift coefficient&Extra Trees Regressor & 0.0257  & 0.0019 & 0.9395 & 14.9734\\
 & Light Gradient Boosting Machine & 0.0357 & 0.0029 & 0.9148 & 	0.4928\\
 & Random Forest Regressor & 0.0253 & 0.0016 & 0.9116 & 0.2552\\
 & Gradient Boosting Regressor & 0.0352  & 0.0031 & 0.8888 & 10.8185\\
\noalign{\smallskip}\hline\noalign{\smallskip}
Computation time&Light Gradient Boosting Machine & 2.4531  & 31.2715 & 0.8731 & 0.1442\\
 & Extra Trees Regressor & 2.4718 & 34.3717 & 0.8524 & 0.1301\\
 & Random Forest Regressor & 2.4533 & 36.7701 & 0.8291 & 0.1306\\
 & Gradient Boosting Regressor & 2.2289 & 31.0430 & 0.8546 & 0.1468\\
\noalign{\smallskip}\hline\noalign{\smallskip}
\end{tabular}
\end{table}

For all output parameters, the hyperparameters in each of the models were finetuned using the Grid Search Cross Validation method -- a systematic method used in machine learning to select the optimal hyperparameters for a model by exhaustively testing all possible combinations within a predefined grid. It involves splitting the dataset into multiple subsets, training the model on different parameter configurations, and evaluating its performance using cross-validation to ensure robustness.  

Subsequently, we employed the three tuned regression models for each output parameter as base regressors within a Voting Regression ensemble model\cite{erdebilli2022ensemble} to attain a more generalized prediction. This involves averaging the individual predictions from the base regressors to yield the final prediction for each target parameter. Table \ref{tab:tuned_models} presents the final base regression models used for each of the output parameters.

\begin{table}[h!]
\centering
\caption{Hyperparameter-tuned base regression models}
\label{tab:tuned_models}
\begin{tabular}{p{2.3cm}p{3.3cm}p{3.3cm}p{3cm}}
\hline\noalign{\smallskip}
Output Parameter & RandomForestRegressor & LGBMRegressor & ExtraTreesRegressor \\
\noalign{\smallskip}\svhline\noalign{\smallskip}
Drag coefficient & (criterion='absolute\_error', max\_depth=8, max\_features=9, n\_estimators=60, random\_state=15, warm\_start=True, n\_jobs=-1) & (min\_child\_samples=30, min\_child\_weight=1e-05, n\_estimators=170, num\_leaves=20, n\_jobs=-1) & (criterion='friedman\_mse', max\_depth=9, max\_features=10, n\_estimators=170, warm\_start=True)\\
Lift coefficient & (max\_depth=5, max\_features=6, n\_estimators=50, random\_state=9, warm\_start=True, n\_jobs=-1) & (boosting\_type='dart', min\_child\_samples=10, min\_child\_weight=1e-05, n\_estimators=190, num\_leaves=20, n\_jobs=-1) & (max\_depth=9, max\_features=10, n\_estimators=140, warm\_start=True)\\
Computation time & (criterion='absolute\_error', max\_depth=3, max\_features=10, n\_estimators=70, random\_state=12, warm\_start=True, n\_jobs=-1) & (min\_child\_weight=1e-05, n\_estimators=50, num\_leaves=20, n\_jobs=-1) & (criterion='absolute\_error', max\_depth=6, max\_features=10, n\_estimators=80)\\
\noalign{\smallskip}\hline\noalign{\smallskip}
\end{tabular}
\end{table}

\begin{figure}[h!]
\centering
\subfloat[]{{\includegraphics[width=\textwidth]{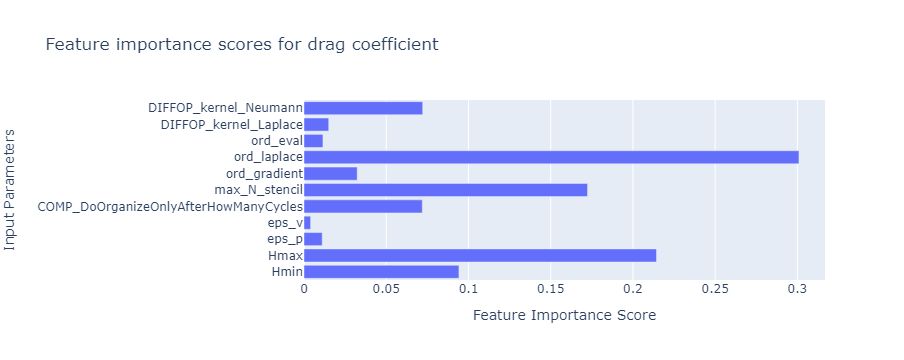} }\label{fig:fi-a}}
\\
\subfloat[]{{\includegraphics[width=\textwidth]{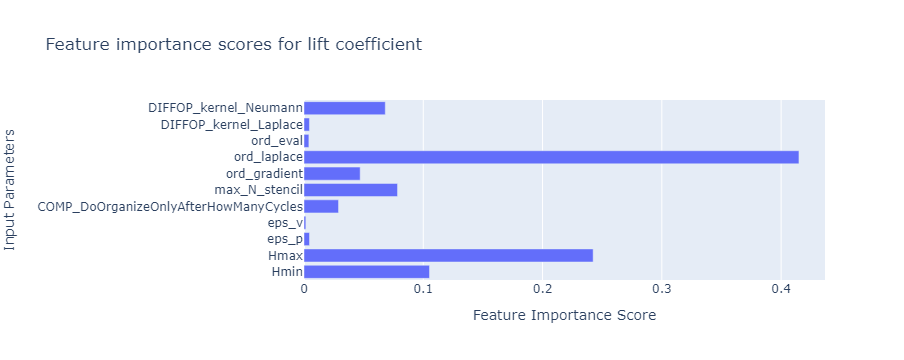} }\label{fig:fi-b}}
\\
\subfloat[]{{\includegraphics[width=\textwidth]{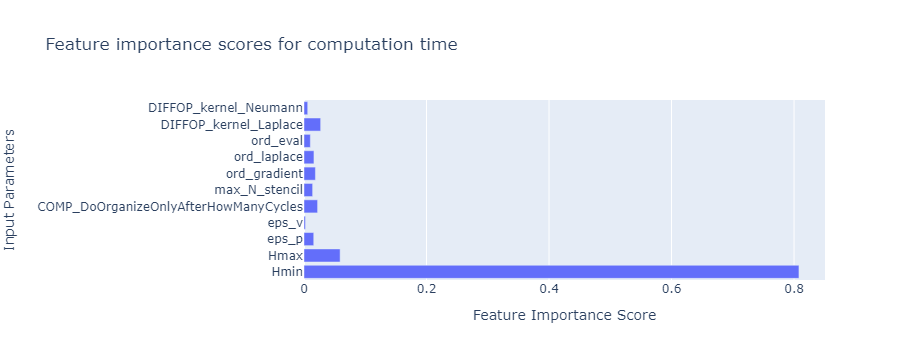} }\label{fig:fi-c}}
\caption{Feature importance scores for (a) drag coefficient, (b) lift coefficient, and (c) computation time}
\label{fig:FI}
\end{figure}

To better understand the behavior of the models, we conducted a Feature Importance analysis, which is illustrated in Figure \ref{fig:FI}. This shed light on the significance of each input parameter in the models for predicting the respective output parameter. Based on this analysis, the dimensionality of the input parameter space for each regression model was reduced to include only those parameters with feature importance score of more than 2\%.  The final set of input parameters considered for each output parameter is outlined in Table \ref{tab:FeatureImportance}. 
\begin{table}[h!]
\centering
\caption{Final set of input parameters after Feature Importance analysis}
\label{tab:FeatureImportance}
\begin{tabular}{p{3.5cm}p{8cm}}
\hline\noalign{\smallskip}
Output Parameter & Input Parameters \\
\noalign{\smallskip}\svhline\noalign{\smallskip}
Drag coefficient & \{\emph{Hmin, Hmax, COMP\_DoOrganizeOnlyAfterHowManyCycles, max\_N\_stencil, ord\_laplace, DIFFOP\_kernel\_Neumann}\}\\
Lift coefficient & \{\emph{Hmin, Hmax, max\_N\_stencil, ord\_gradient, ord\_laplace, \linebreak DIFFOP\_kernel\_Neumann}\}\\
Computation time & \{\emph{Hmin, Hmax, DIFFOP\_kernel\_Laplace}\}\\
\noalign{\smallskip}\hline\noalign{\smallskip}
\end{tabular}
\end{table}

The final regression models were then used to predict intervals within which the output parameters might lie, given the high uncertainty associated with point estimates (specific values for outputs that are predicted by regression models). This approach acknowledges and quantifies the uncertainty inherent in the predictions, enabling more informed decision-making. To achieve this, the method of Conformal Prediction was implemented. It is a statistical tool used in regression analysis to estimate prediction intervals for ML models. Furthermore, it provides a framework that is model agnostic, allowing it to adapt to various regression algorithms without relying on specific assumptions about the data distribution or model form, which enhances its robustness and applicability. 

\begin{figure}[h!]
    \centering
    \includegraphics[width=\textwidth]{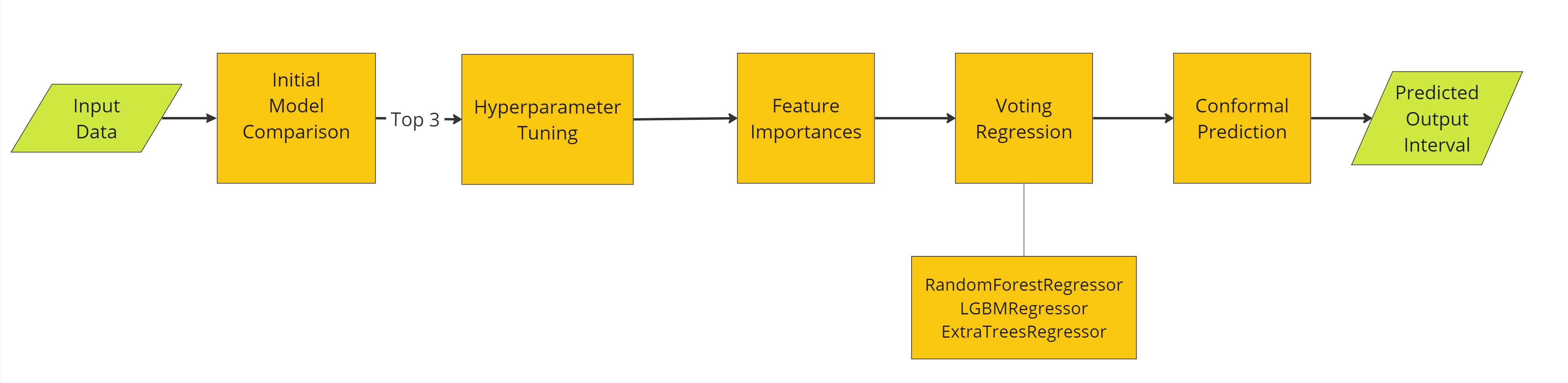}
    \caption{Flow diagram for each output parameter interval prediction}
    \label{fig:predflow}
\end{figure}

In this implementation, the tolerance parameter $\alpha$ plays a significant role in balancing the trade-off between the size of the predicted intervals and the quality of the prediction. After discussing with domain experts, we have used $\alpha_{C_\mathrm{D}} = 0.15$, $\alpha_{C_\mathrm{L}} = 0.2$, and $\alpha_{\mathrm{CT}} = 0.1$ for the models predicting drag coefficient, lift coefficient, and computation time (CT) intervals, respectively. This means that the model may have 15\%, 20\%, and 10\%, respectively, of the actual values outside the predicted intervals. Figure \ref{fig:predflow} summarizes the flow of information for the prediction of each output parameter as described in this subsection.

\section{Results and Key Findings}
\label{sec:Results}

In this section, we present the results of our investigation of the influence of various parameters on simulation results quality and computation time using the MESHFREE software. Our findings provide valuable guidance for optimizing parameter selection as well as enhancing the efficiency and accuracy of fluid flow simulations.

\subsection{Active Learning Module}
\label{sec:ResultsAL}
\begin{figure}[h!]
    \centering
    \includegraphics[width=\textwidth]{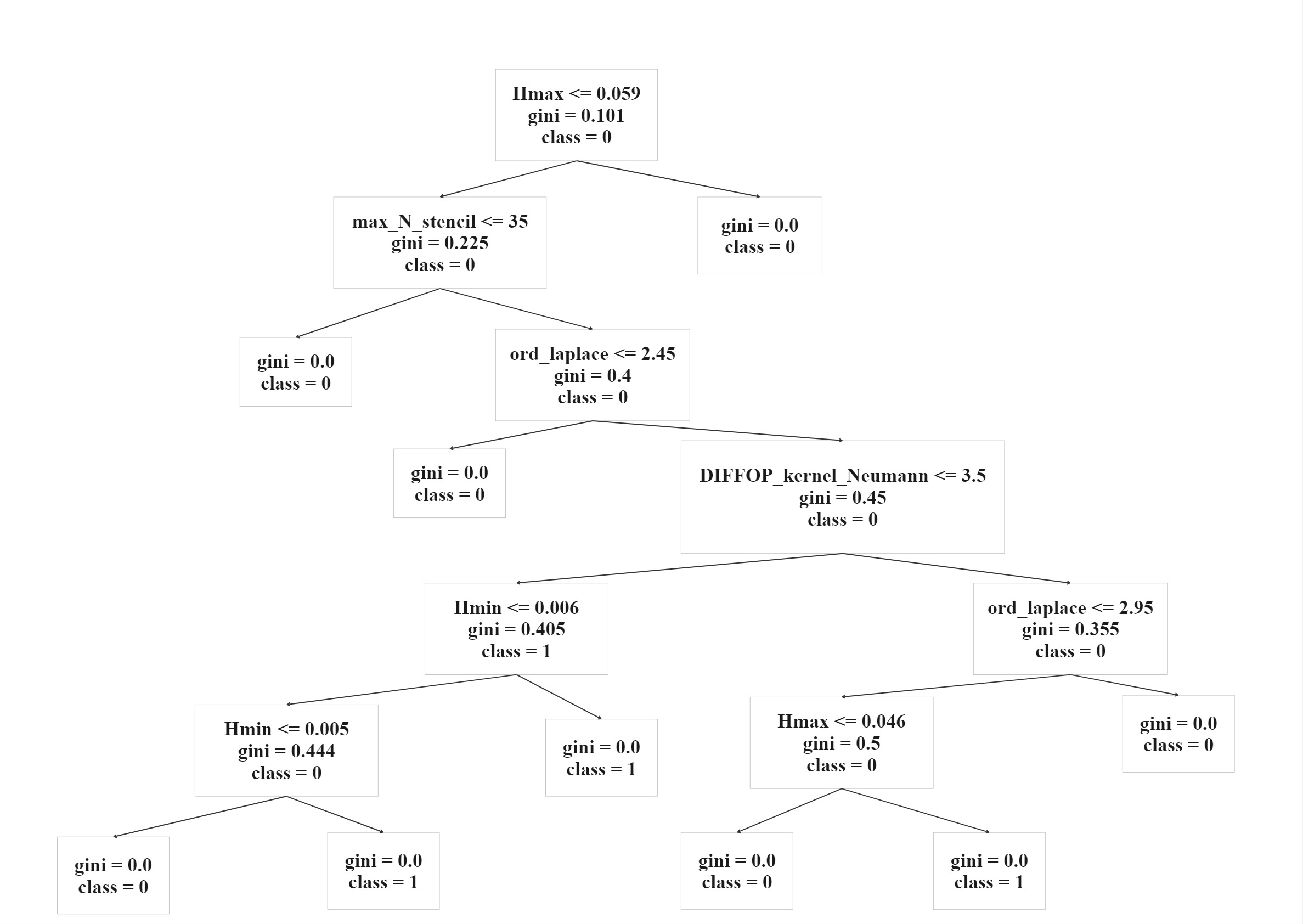}
    \caption{Final decision-tree for selecting optimized parameter ranges}
    \label{fig:ALDecisionTree}
\end{figure}

A representation of the final decision-tree from the Active Learning module explained in Section \ref{sec:AL} is illustrated in Figure \ref{fig:ALDecisionTree}. \emph{class 1} and \emph{class 0} in the leaf nodes refer to cases with optimal and non-optimal outputs, respectively. Our findings regarding the optimized input parameter ranges are as follows:
\begin{itemize}
    \item \emph{Hmin} should be kept between (ideally) \emph{0.005} and \emph{0.007}.
    \item \emph{Hmax} should be kept between (ideally) \emph{0.046} and \emph{0.06} with \emph{Hmax} $>$ \emph{Hmin}.
    \item \emph{COMP\_DoOrganizeOnlyAfterHowManyCycles} can be chosen according to the initial range. However, to keep the computation time to a minimum, a value of \emph{3} or \emph{4} was found to be ideal.
    \item Best results are achieved with \emph{max\_N\_stencil} = \emph{40}.
    \item \emph{eps\_p} and \emph{eps\_v} can be chosen according to the initial range, too. But using a higher value is recommended to keep the computation time to a minimum.
    \item \emph{ord\_grad} can be chosen according to the initial range.
    \item If \emph{Hmax} $\leq$ \emph{0.046}, \emph{ord\_laplace} = \emph{2.9} provides the best results, otherwise \emph{3}.
    \item \emph{ord\_eval} can be chosen according to the initial range.
    \item \emph{DIFFOP\_kernel\_Laplace} can be chosen according to the initial range.
    \item \emph{DIFFOP\_kernel\_Neumann} can be chosen according to the initial range, too. However, in combination with \emph{ord\_laplace} = \emph{2.9} and \emph{Hmax} $\leq$ 0.046 a value of \emph{5} is recommended.
\end{itemize}

Maintaining these optimized parameter ranges, the user will have an 87\% chance of achieving a good trade-off between accuracy and efficiency.

\subsection{Interval Prediction}
\begin{table}
\centering
\caption{Error metrics of the final regression models}
\label{tab:FinalModelPerf}
\begin{tabular}{p{4.5cm}p{1.3cm}p{1.3cm}p{1.3cm}p{1cm}}
\hline\noalign{\smallskip}
Model & MAE & MSE & R2 & MAPE  \\
\noalign{\smallskip}\svhline\noalign{\smallskip}
Drag coefficient model & 0.4661 & 0.4012 & 0.9879 & 0.1871\\
Lift coefficient model & 0.0134 & 0.0005 & 0.9812 & 3.3515\\
Computation time model & 1.4157 & 16.1815 & 0.9313 & 0.0906\\
\noalign{\smallskip}\hline\noalign{\smallskip}
\end{tabular}
\end{table}
The error metrics of the final ensemble models are shown in Table \ref{tab:FinalModelPerf}. A clear improvement in the performance of the individual models can be observed, as compared to Table \ref{tab:InitPerfComp}. Tab 2 of the dashboard visualizes the performance of our trained regression models on a test dataset of 20 random test cases. Here is a summary of our observations:
\begin{itemize}
    \item A higher value for \emph{COMP DoOrganizeOnlyAfterHowManyCycles} leads to predictions with lower variance in the drag coefficient, compared to lower values.
    \item For \emph{max\_N\_stencil} = \emph{30}, the actual values of the lift coefficient are more often outside the upper and lower limits of the predicted interval, indicating more incorrect predictions than for \emph{max\_N\_stencil} = \emph{40}.
    \item The actual outputs for cases with \emph{ord\_laplace} = \emph{2.9} lie perfectly in between the predicted interval for both the drag coefficient and the lift coefficient.
\end{itemize}

The amount of stochasticity in the simulation due to specific feature parameter values or a combination of them play a significant role in whether our predicted intervals are able to accurately capture the actual output within their range. Cases with \emph{max\_N\_stencil} = \emph{40} but higher values for \emph{Hmin} as well as \emph{Hmax} are more likely to be incorrectly predicted than others. While a higher value of \emph{COMP\_DoOrganizeOnlyAfterHowManyCycles} also brings in its fair share of randomness in the simulation, keeping a stencil size of \emph{max\_N\_stencil} = \emph{40} was observed to neutralize its effect on the predictions, reducing the corresponding computation time to a minimum. 

\begin{figure}[h!]
    \centering
    \includegraphics[width=\textwidth]{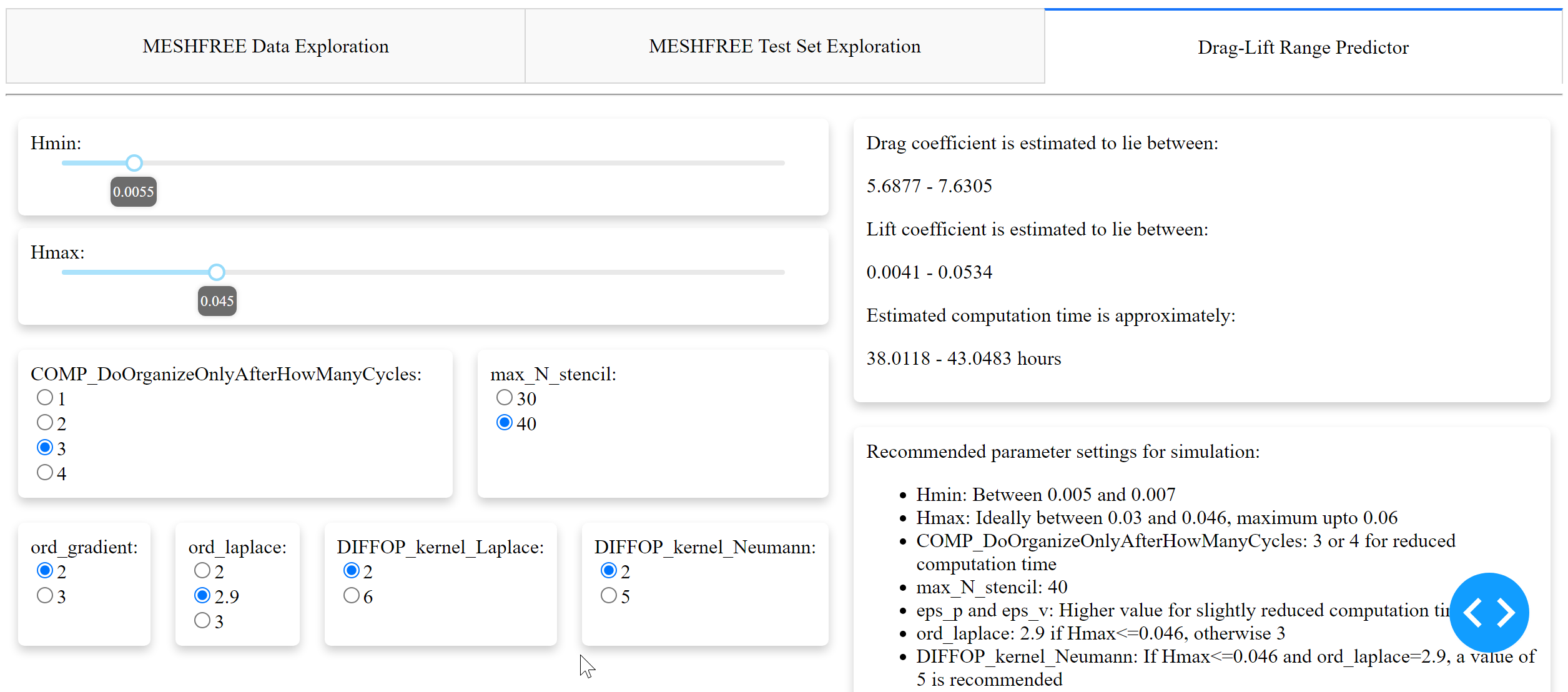}
    \caption{Tab 3 of dashboard showing the implemented tool for predictions}
    \label{fig:Tab3Dashboard}
\end{figure}

We have implemented a predictor tool using our trained models in Tab 3 of the dashboard, as shown in Figure \ref{fig:Tab3Dashboard}. It helps users get an estimate on the ranges for the simulation quality and computation time upon adjusting the input parameter values. Parameters \emph{eps\_p}, \emph{eps\_v} and \emph{ord\_eval} have been excluded from this tool since they were not included in the predictions for any of the output parameter intervals.

\section{Conclusion and Outlook}
\label{sec:Conclusion}

In this publication, we delved into the realm of meshfree simulation methods, particularly focusing on the integration of machine learning into the MESHFREE software developed by Fraunhofer ITWM and Fraunhofer SCAI. By moving away from conventional mesh-based approaches, this software streamlines the simulation process by eliminating the need for extensive mesh generation and re-meshing, especially in cases of complex and deformable domains or free surface flows.

During the investigation, we identified key parameters inherent in the MESHFREE software stemming from the underlying GFDM. Understanding these parameters is pivotal for configuring simulations effectively, particularly for inexperienced users. Our primary goal was to empower users with the ability to navigate the simulation tool seamlessly, even in the absence of prior knowledge of meshfree methods and the specific method. To achieve this goal, we have chosen a simple test case, the 3D flow around a cylinder according to \cite{Schaefer1996}.

Our research emphasizes the importance of parameter selection for the balance between simulation accuracy and computational cost. We addressed the challenge of parameter interdependence and fine-tuning by providing pre-selected parameter ranges tailored for optimal efficiency and accuracy, using active learning strategies. By systematically recording the effects of parameters on computation time, drag and lift coefficients, and using ensemble decision-tree-based regression models, we facilitated informed decision-making and optimized computational resources.

Future research efforts will explore advanced parameter optimization and sensitivity analysis techniques to further improve the efficiency and reliability of meshfree simulations. In addition, efforts should be directed towards the development of user-friendly interfaces and tutorials to promote the acceptance of meshfree simulation tools among an even broader user base than today.

In essence, this study contributes to the continuous development of meshfree simulation methods by providing insights and solutions to the challenges encountered in effectively configuring and running simulations. By empowering users with the necessary tools and knowledge, we are paving the way for a more accessible and efficient approach to simulating fluid flows and more generally continuum mechanical processes in various engineering applications.

\begin{acknowledgement}
This research was partially funded by the Hessian Ministry of Higher Education, Research, Science and the Arts, Germany (FL1, Mittelbau).
\end{acknowledgement}

\section*{Author Contribution}
The authors contributed as follows: PB contributed to conceptualization, data generation, methodology, visualization, results analysis and interpretation, as well as writing the original draft, and reviewing and editing the manuscript. MP was involved in conceptualization, use case design, data generation, visualization, results analysis and interpretation, writing the original draft, and reviewing and editing. CS contributed to conceptualization, use case design, data generation, and reviewing and editing. IM contributed to conceptualization, data generation, supervision, and reviewing and editing. SG contributed to conceptualization, supervision, and reviewing and editing.

All authors reviewed the results and approved the final version of the manuscript. 

\section*{Conflict of Interest}
The authors declare that there are no conflicts of interest associated with this work.

%

%
%
%


\end{document}